\documentclass[10pt, letter, twocolumn]{article} 

\usepackage[english]{babel} 

\usepackage{amsmath,amsfonts,amsthm} 

\usepackage[svgnames]{xcolor} 
\usepackage{mathtools}
\usepackage{array}
\usepackage{hyperref}

\usepackage{color}

\usepackage{flushend}

\definecolor{boxblue}{RGB}{234,249,254}
\definecolor{lineblue}{RGB}{141,214,239}

\definecolor{light}{rgb}{0.4, 0.4, 0.4}
\definecolor{lblue}{rgb}{0.56,0.80,0.92}
\definecolor{llblue}{rgb}{0.9, 0.9, 0.9}
\definecolor{dgrey}{rgb}{0.2, 0.2, 0.2}
\definecolor{lgrey}{rgb}{0.5, 0.5, 0.5}
\def\light#1{{\color{light}#1}}

\usepackage[footnotesize, up, textfont=sf]{caption} 

\DeclareCaptionFormat{figfont}{{\color{lblue}\usefont{OT1}{ugq}{n}{n}{#1. \ }}#3}
\usepackage{booktabs} 
\usepackage{lastpage} 

\usepackage{graphicx,dblfloatfix} 
\usepackage{tikz}
\usetikzlibrary{backgrounds,calc}
\usepackage{enumitem} 
\usepackage[framemethod=TikZ]{mdframed}

\newsavebox{\picbox}

\setlist{noitemsep} 
\usepackage{todonotes}

\usepackage{sectsty} 
\sectionfont{
    \Large\color{lblue}\usefont{OT1}{ugq}{n}{n}
}

\subsectionfont{
    \color{light}\usefont{OT1}{ugq}{n}{n}
}

\hypersetup{
    colorlinks=true,
    linkcolor=black,
    urlcolor=black,
    citecolor=black
}

\usepackage{geometry} 

\usepackage[T1]{fontenc} 
\usepackage[utf8]{inputenc} 

\usepackage{anyfontsize}

\usepackage{fancyhdr} 
\fancypagestyle{firstpage}{ 
	\fancyhf{}
    \lfoot{
        \scriptsize
        \light{
            {\fontfamily{cmss}\selectfont
            This article has been accepted for publication in IEEE Robotics \& Automation Magazine \\ Published Version Available at \url{https://doi.org/10.1109/MRA.2023.3270221}
            }
        }
    }
    \rfoot{\footnotesize
    \fontfamily{cmss}\selectfont
    \light{\thepage}\
    } 
}

\usepackage[firstpage=true]{background}

\backgroundsetup{
scale=1,
angle=0,
opacity=1,
contents={\begin{tikzpicture}[remember picture,overlay]
 \path [top color = white, bottom color = llblue] ($(current page.north west)-(4in,4in)$) rectangle (current page.south east);
\end{tikzpicture}}
}

\newcommand{\authorstyle}[1]{\color{dgrey}\centering{\usefont{OT1}{phv}{n}{n}#1}} 

\newcommand{\titlestyle}[1]{\color{dgrey}\fontsize{35}{45}\centering{\usefont{OT1}{ugq}{n}{n}#1}}

\usepackage{titling} 

\pretitle{
	\vspace{120pt} 
	\fontsize{40}{46}\usefont{OT1}{phv}{b}{n}\selectfont 
}

\posttitle{\par\vskip 25pt} 

\preauthor{} 

\postauthor{ 
	\vspace{-15pt} 
}

\usepackage{lettrine} 
\newcommand{\initial}[1]{ 
	\lettrine[lines=4,findent=5pt,nindent=0pt]{
		\color{lblue}
		{#1}
	}{}%
}

\usepackage[backend=biber,style=ieee,natbib=true,doi=false,isbn=false,url=false,eprint=false]{biblatex} 

\AtNextBibliography{\footnotesize}
\usepackage[autostyle=true]{csquotes} 

\usepackage{wrapfig}

\usepackage{bbding}

\usepackage{amsmath}
\usepackage{amssymb}
\usepackage{accents}
\usepackage{bm}
\usepackage{xifthen}
\usepackage{color}
\usepackage{fancyvrb}
\usepackage{graphicx,scalerel}

\newcommand{\ba}{\begin{eqnarray}}
\newcommand{\ea}{\end{eqnarray}}

\newcommand{\unit}[1]{\mbox{$\mathrm{\,#1}$}}

\renewcommand{\deg}{\mbox{${}^\circ$}}

\newcommand{\R}{\mathbb{R}}

\newcommand{\presup}[1]{\,{}^{\scriptscriptstyle #1}\!}

\newcommand{\pose}[1][_NONE]{\ifthenelse{\equal{#1}{_NONE}}{}{\presup{#1}}{\mathbf{\xi}}}
\newcommand{\estpose}[1][_NONE]{\ifthenelse{\equal{#1}{_NONE}}{}{\presup{#1}}{\hat{\mathbf{\xi}}}}
\newcommand{\hpose}[1][_NONE]{\ifthenelse{\equal{#1}{_NONE}}{}{\presup{#1}}{\hat{\mathbf{\xi}}}}
\newcommand{\posedot}[1][_NONE]{\ifthenelse{\equal{#1}{_NONE}}{}{\presup{#1}}{\mathbf{\nu}}}
\newcommand{\q}[1][_NONE]{\ifthenelse{\equal{#1}{_NONE}}{}{\presup{#1}}{\mathring{q}}}
\newcommand{\uquat}[2][_NONE]{\ifthenelse{\equal{#1}{_NONE}}{}{\presup{#1}}{\mathring{#2}}}
\newcommand{\duquat}[2][_NONE]{\ifthenelse{\equal{#1}{_NONE}}{}{\presup{#1}}{\dot{\mathring{#2}}}}
\newcommand{\quat}[2][_NONE]{\ifthenelse{\equal{#1}{_NONE}}{}{\presup{#1}}{\breve{#2}}}
\DeclareMathAlphabet{\mathitbf}{OML}{cmm}{b}{it}
\newcommand{\twist}[2][_NONE]{\ifthenelse{\equal{#1}{_NONE}}{}{\presup{#1}}{\mathcal{S}}}
\renewcommand{\vec}[2][_NONE]{\ifthenelse{\equal{#1}{_NONE}}{}{\presup{#1}}{\boldsymbol #2}}
\newcommand{\hvec}[2][_NONE]{\ifthenelse{\equal{#1}{_NONE}}{}{\presup{#1}}{\tilde{\vec{#2}}}}
\newcommand{\evec}[2][_NONE]{\ifthenelse{\equal{#1}{_NONE}}{}{\presup{#1}}{\hat{\vec{#2}}}}
\newcommand{\bvec}[2][_NONE]{\ifthenelse{\equal{#1}{_NONE}}{}{\presup{#1}}{\bar{\vec{#2}}}}
\newcommand{\dhvec}[2][_NONE]{\ifthenelse{\equal{#1}{_NONE}}{}{\presup{#1}}{\dot{\tilde{\vec{#2}}}}}
\newcommand{\dvec}[2][_NONE]{\ifthenelse{\equal{#1}{_NONE}}{}{\presup{#1}}{\dot{\vec{#2}}}}
\newcommand{\ddvec}[2][_NONE]{\ifthenelse{\equal{#1}{_NONE}}{}{\presup{#1}}{\ddot{\vec{#2}}}}

\newcommand{\mat}[2][_NONE]{\ifthenelse{\equal{#1}{_NONE}}{}{\presup{#1}\,}{{\mathbf #2}}}
\newcommand{\dmat}[2][_NONE]{\ifthenelse{\equal{#1}{_NONE}}{}{\presup{#1}\,}{{\dot{\mathbf #2}}}}
\newcommand{\emat}[2][_NONE]{\ifthenelse{\equal{#1}{_NONE}}{}{\presup{#1}\,}{\hat{\mathbf#2}}}
\newcommand{\hmat}[2][_NONE]{\ifthenelse{\equal{#1}{_NONE}}{}{\presup{#1}\,}{\tilde{\mathbf{#2}}}}
\newcommand{\matfn}[3][_NONE]{\ifthenelse{\equal{#1}{_NONE}}{}{\presup{#1}}{{\mat{#2}}\left(#3\right)}}
\newcommand{\Rt}[2][_NONE]{\ifthenelse{\equal{#1}{_NONE}}{}{\presup{#1}}{{\bf R}\left(#2\right)}}

\newcommand{\point}[2][_NONE]{\ifthenelse{\equal{#1}{_NONE}}{}{\ensuremath{\presup{#1}}}{\ensuremath{\mathrm{#2}}}}

\newfont{\School}{pncr}
\newfont{\eightTR}{pncr at 8pt}

\fvset{formatcom={\color{blue}\baselineskip=-\maxdimen\lineskip=0pt},xleftmargin=4mm,commentchar=!}
\DefineVerbatimEnvironment{Code}{Verbatim}{}
\DefineVerbatimEnvironment{CodeSmall}{Verbatim}{fontsize=\scriptsize,xleftmargin=1mm,commentchar=!}
\DefineVerbatimEnvironment{CodeNum}{Verbatim}{numbers=left,numbersep=4pt,fontsize=\footnotesize,xleftmargin=4mm}
\newcommand{\func}[2][_NONE]{\ifthenelse{\equal{#1}{_NONE}}{\index{code}{#2}}{\index{code}{#1}}\ifthenelse{\boolean{draft}}{{\color{green}\Verb+#2+}}{\Verb+#2+}}
\newcommand{\methodb}[2]{\index{code}{#1@\textbf{#1}!.#2}\ifthenelse{\boolean{draft}}{{\color{magenta}\Verb+#1.#2+}}{\Verb+#1.#2+}}
\newcommand{\method}[2]{\index{code}{#1@\textbf{#1}!.#2}\ifthenelse{\boolean{draft}}{{\color{magenta}\Verb+#2+}}{\Verb+#2+}}
\newcommand{\class}[1]{\index{code}{#1@\textbf{#1}}\ifthenelse{\boolean{draft}}{{\color{cyan}\Verb+#1+}}{\Verb+#1+}}
\newcommand{\property}[1]{\index{property}{#1}\ifthenelse{\boolean{draft}}{{\color{cyan}\Verb+#1+}}{\Verb+#1+}}

\newcommand{\SE}[1]{\ensuremath{\mathrm{{\bf SE}(#1)}}}

\newcommand{\iskx}[1]{\vee_{\times}\left( #1\right)}
\newcommand{\skx}[1]{\left[#1\right]_{\times}}
\newcommand{\sk}[1]{\left[#1\right]}

\usepackage{xparse}
\NewDocumentCommand{\ovec}{ o m o }{%
  \IfValueT{#1}{\presup{#1}}%
  \vec{#2}%
  \IfValueT{#3}{_{#3}}
  ^{\scalebox{0.6}{\#}}%
  }
\NewDocumentCommand{\omat}{ o m o }{%
  \IfValueT{#1}{\presup{#1}}%
  \mat{#2}%
  \IfValueT{#3}{_{#3}}
  ^{\scalebox{0.6}{\#}}%
  }
\NewDocumentCommand{\obspose}{ o o }{%
  \IfValueT{#1}{\presup{#1}}%
  \mathbf{\xi}%
  \IfValueT{#2}{_{#2}}
  ^{\scalebox{0.6}{\#}}%
  }

\renewcommand{\q}{\vec{q}}

\newcommand{\qd}{\dot{\vec{q}}}
\newcommand{\qdd}{\ddot{\vec{q}}}
\newcommand{\J}{\mat{J}}

\newcommand{\Jd}{\mathrlap{ \dot{\phantom{\ \mat{J}}} }{\mat{J}}}

\newcommand{\JR}{\mat{J_{\!R}}}
\renewcommand{\R}{\mat{R}}

\newcommand{\T}{\mat{T}}

\renewcommand{\t}{\vec{t}}

\newcommand{\Jv}{\mat{J}_\nu}
\newcommand{\Jvx}[1]{\mat{J}_{{\!\nu}_{#1}}}
\newcommand{\Jw}{\mat{J_{\!\omega}}}
\newcommand{\Jwx}[1]{\mat{J}_{{\!\omega}_{#1}}}
\renewcommand{\H}{\mat{H}}

\newcommand{\Ht}{\mat{H_{\!t}}}
\newcommand{\HR}{\mat{H_{\!R}}}

\newcommand{\Hwx}[1]{\mat{H}_{{\!\alpha}_{#1}}}
\newcommand{\Hv}{\mat{H}_{\!a}}
\newcommand{\Hvx}[1]{\mat{H}_{{\!a}_{#1}}}
\newcommand{\TR}[1]{\mat{T}_{{\!\R}_{#1}}}
\newcommand{\Tt}[1]{\mat{T}_{{\!\t}_{#1}}}

\usepackage{newfloat}
\DeclareFloatingEnvironment[
    fileext=los,
    listname={List of Excurses},
    name=Excurse,
    placement=tbhp,
    within=section,
]{excurse}

\DeclareCaptionFormat{excfont}{
    \normalsize{
        \color{lblue}
        \usefont{OT1}{ugq}{n}{n}{
            #1. \
        }
        \color{dgrey} #3
    }
}
\title{
\vspace{-148pt}
\protect\centering \protect\includegraphics[width=0.35\linewidth]{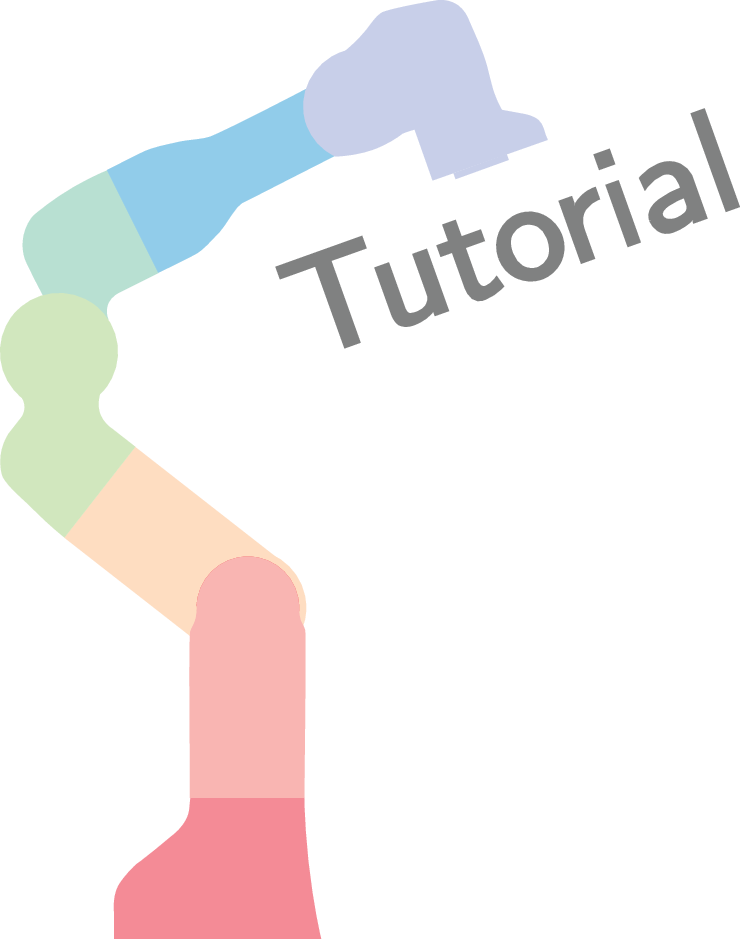}\\
\vspace{32pt}
\Huge
\titlestyle{Manipulator \\\vspace{5pt}Differential Kinematics}\\
\vspace{10pt}
\protect\centering\Large Part 2: Acceleration and Advanced Applications\\
[0.2cm]
}

\author{
    {
       \authorstyle{By Jesse Haviland and Peter Corke}\\[0.5cm]
    }
}

\date{}

\begin{document}

\maketitle

\thispagestyle{firstpage}


\begin{figure*}[b!] 
    \newcounter{mytempeqncnt}
    \normalsize 
    \setcounter{mytempeqncnt}{\value{equation}}
    \setcounter{equation}{5} 
    
    \vspace*{4pt} 
    \hrulefill
    \begin{align} \label{eq:ddets}
        \frac{\partial^2 \T}
            {\partial q_k q_j} &= 
        \begin{pmatrix} 
            \HR_{j_k} & \Ht_{j_k} \\ 
            0  & 0 
        \end{pmatrix} \nonumber \\
        &= 
        \frac{\partial} 
            {\partial q_k} 
        \left( 
        \prod_{i=1}^{\mu(j)-1} \mat{E}_i(\eta_i)
        \
        \frac{\mathrm{d} \mat{E}_{\mu(j)}(q_j)} 
            {d q_j} 
        \
        \prod_{i=\mu(j)+1}^{M} \mat{E}_i(\eta_i)  
        \right) \nonumber  \\
        &= \left\{ 
        \begin{matrix*}[l]
            \prod\limits_{i=1}^{\mu(k) - 1} \mat{E}_i(\eta_i)
            \
            \frac{\mathrm{d} \mat{E}_{\mu(k)}(q_k)}
                {d q_k}
            \
            \prod\limits_{i=\mu(k)+1}^{\mu(j) - 1} \mat{E}_i(\eta_i) 
            \
            \frac{\mathrm{d} \mat{E}_{\mu(j)}(q_j)}
                {d q_j} 
            \
            \prod\limits_{i=\mu(j)+1}^{M} \mat{E}_i(\eta_i)
            & \mbox{if} \ k < j \\
            \prod\limits_{i=1}^{\mu(k) - 1} \mat{E}_i(\eta_i) 
            \
            \frac{\mathrm{d}^2 \mat{E}_{\mu(k)}(q_k)}
                {d q_k^2} 
            \
            \prod\limits_{i=\mu(k)+1}^{M} \mat{E}_i(\eta_i) 
            & \mbox{if} \ k = j \\
            \prod\limits_{i=1}^{\mu(j) - 1} \mat{E}_i(\eta_i) 
            \
            \frac{\mathrm{d} \mat{E}_{\mu(j)}(q_j)}
                {d q_j}
            \
            \prod\limits_{i=\mu(j)+1}^{\mu(k) - 1} \mat{E}_i(\eta_i) 
            \
            \frac{\mathrm{d} \mat{E}_{\mu(k)}(q_k)}
                {d q_k} 
            \
            \prod\limits_{i=\mu(k)+1}^{M} \mat{E}_i(\eta_i)
            & \mbox{if} \ k > j \\
        \end{matrix*}
        \right.
    \end{align}
    \setcounter{equation}{\value{mytempeqncnt}} 
\end{figure*}
    

\initial{K}inematics, in the context of robotic manipulators is concerned with the relationship
between the position of the robot's joints and the pose of its end effector, as well as the relationships between various derivatives of those quantities.
In this second part of our two-part tutorial, we focus on second-order differential kinematics and subsequent applications. These applications demonstrate advanced techniques which are highly relevant to topics including sensor-based control, constrained control, and motion planning.

We start by introducing the second-order differential kinematics and the manipulator Hessian. The second-order differential kinematics expose a relationship between the robot's joint velocities and the end-effector acceleration. We then describe the differential kinematics' analytical forms, which are essential to dynamics applications. Subsequently, we provide a general formula for higher-order derivatives before detailing and experimenting with three advanced applications.

The first application we consider is advanced velocity control -- an important topic for reactive and sensor-based control tasks. We demonstrate this by extending resolved-rate motion control (RRMC) to perform additional sub-tasks while still achieving its goal and then reformulate the problem as a quadratic program to enable greater flexibility and additional constraints.
We then take another look at numerical inverse kinematics (IK) with an emphasis on adding constraints to improve robustness and solvability.
We subsequently present a comprehensive experiment that compares the performance and characteristics of each IK method on three different manipulators.
Finally, we analyse how incorporating the manipulator Hessian into a motion controller can help to escape singularities.

In Part 1 \cite{dkt1}, we described a method of modelling kinematics using the elementary transform sequence (ETS), before formulating forward kinematics and the manipulator Jacobian. We then described some introductory but fundamental applications of the manipulator Jacobian including RRMC, numerical IK, and some manipulator performance measures.  Part 1 provides many important definitions, functions and conventions, and we recommend that readers review Part 1 before reading this Part.

Once again, we have provided Jupyter Notebooks to accompany each section within this tutorial. The Notebooks are written in Python and use the Robotics Toolbox for Python, and the Swift Simulator \cite{rtb} to provide full implementations of each concept, equation, and algorithm presented in this tutorial. The Notebooks use rich Markdown text and \LaTeX\ equations to document and communicate key concepts. While not absolutely essential, for the most engaging and informative experience, we recommend working through the Jupyter Notebooks while reading this article. The Notebooks and setup instructions can be accessed at \href{https://github.com/jhavl/dkt}{github.com/jhavl/dkt}.


\section*{Deriving the Manipulator Hessian}
\subsection*{The Manipulator Hessian} \label{sec:ddH}

We begin with the forward kinematics of a manipulator as described in Part 1
\begin{align} \label{eq:ets1}
    {^0\T_e(t)}  &= {\cal K}(\q(t)) \nonumber\\
    &= \prod_{i=1}^{M} \mat{E}_i(\eta_i).
 \end{align}
where $\q(t) \in \mathbb{R}^n$ is the vector of joint generalised coordinates, $n$ is the number of joints, and $M$ is the number of elementary transforms $\mat{E}_i \in \SE{3}$. From the derivative of (\ref{eq:ets1}) we can express the spatial translational and angular velocity as a function of the joint coordinates and velocities
\begin{align} \label{eq:e0}
    \vec{\nu} =
    \begin{pmatrix}
        \vec{v} \\ \vec{\omega}
    \end{pmatrix}
    &=
    \mat{J}(\q)\qd
\end{align}
where $\mat{J}(\q)$ is the manipulator Jacobian, $\vec{v} = (v_x, \ v_y, \ v_z)$, and $\vec{\omega} =(\omega_x, \ \omega_y, \ \omega_z)$. Taking the temporal derivative gives
\begin{align}
    \begin{pmatrix}
        \dvec{v} \\ \dvec{\omega}
    \end{pmatrix}
    &=
    \begin{pmatrix}
        \vec{a} \\ \vec{\alpha}
    \end{pmatrix}
    =
    \Jd \qd + \J \qdd
\end{align}
where $\vec{a} \in \mathbb{R}^3$ is the end-effector translational acceleration, $\vec{\alpha} \in \mathbb{R}^3$ is the end-effector angular acceleration, and
\begin{align} \label{eq:a1}
    \Jd
    &=
    \dfrac{\mathrm{d} \mat{J}(\q)}
        {\mathrm{d} t} \nonumber \\
    &=
    \dfrac{\partial \mat{J}(\q)}
        {\partial q_1} \dot{q}_1
    +
    \dfrac{\partial \mat{J}(\q)}
        {\partial q_2} \dot{q}_2
    + \cdots +
    \dfrac{\partial \mat{J}}
        {\partial q_n} \dot{q}_n \nonumber \\
    &=
    \begin{pmatrix}
        \dfrac{\partial \mat{J}(\q)}
            {\partial q_1} &
        \dfrac{\partial \mat{J}(\q)}
            {\partial q_2} &
        \cdots &
        \dfrac{\partial \mat{J}(\q)}
            {\partial q_n}
    \end{pmatrix}
    \qd \nonumber \\
    &=
    \mat{H}(\vec{q}) \qd \nonumber \\
    &=
    \begin{pmatrix}
        \mat{H}_a(\vec{q}) \\ \mat{H}_\alpha(\vec{q})
    \end{pmatrix} \qd
\end{align}
where $\mat{H}(\vec{q}) \in \mathbb{R}^{n \times 6 \times n}$ is the manipulator Hessian tensor, which is the partial derivative of the manipulator Jacobian with respect to the joint coordinates, $\mat{H}_a(\vec{q}) \in \mathbb{R}^{n  \times 3 \times n}$ forms the translational component of the Hessian, and $\mat{H}_\alpha(\vec{q}) \in \mathbb{R}^{n \times 3 \times n}$ forms the angular component of the Hessian.

\begin{figure*}[b]
    \centering
    \includegraphics[height=5.3cm]{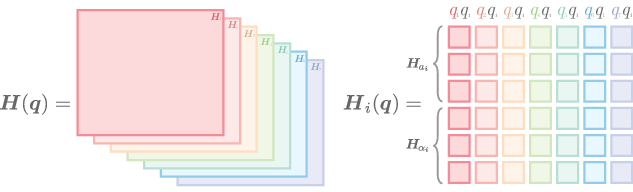}
    \caption{
       Visualisation of the Hessian $\H(\q)$ representing a 7-joint manipulator. Each slice of the Hessian $\H_i(\q)$ represents the acceleration of the end-effector caused by the velocities of each joint $\q$ with respect to the velocity of joint $q_i$.
       Within a slice,
       the top three rows $\H_{\alpha_i}$ correspond to the linear acceleration,
       while the bottom three rows $\H_{a_i}$ correspond to the angular acceleration,
       of the end-effector $\vec{\alpha}$ caused by the velocities of the joints.
    }
    \label{fig:hessian}
\end{figure*}

The second partial derivative of a pose with respect to the joint variables $q_j$ and $q_k$,
can be obtained by taking the derivative of
\begin{align} \label{eq:pdr}
    \frac{\partial \matfn{T}{\vec{q}}}
        {\partial q_j} 
    &=
    \frac{\partial} 
        {\partial q_j} \nonumber 
    \left(
        \mat{E}_1(\eta_1) \mat{E}_2(\eta_2) \cdots \mat{E}_M(\eta_M) 
    \right)\\
    &= 
    \prod_{i=1}^{\mu(j)-1} \mat{E}_i(\eta_i) 
    \frac{\mathrm{d} \mat{E}_{\mu(j)}(q_j)} 
        {\mathrm{d} q_j} 
    \prod_{i=\mu(j)+1}^{M} \mat{E}_i(\eta_i).
\end{align}
with respect to $q_k$. This results in (\ref{eq:ddets}), where the function $\mu(j)$ returns the index in (\ref{eq:ets1}) where $q_j$ appears as a variable.

In (\ref{eq:ddets}), the derivative of an elementary transform with respect to a joint coordinate is obtained using one of
\addtocounter{equation}{1}
\begin{align}
    \dfrac{\mathrm{d} \TR{x}(\theta)}
          {\mathrm{d} \theta}
    &=     
    \begin{pmatrix}
        0 & 0 & 0 & 0 \\
        0 & 0 & -1 & 0 \\
        0 & 1 & 0 & 0 \\
        0 & 0 & 0 & 0 
    \end{pmatrix} \TR{x}(\theta) = \big[ \evec{R}_x \big] \TR{x}(\theta), \label{eq:ERx} \\
    \dfrac{\mathrm{d} \TR{y}(\theta)}
          {\mathrm{d} \theta}
    &=     
    \begin{pmatrix}
        0 & 0 & 1 & 0 \\
        0 & 0 & 0 & 0 \\
        -1 & 0 & 0 & 0 \\
        0 & 0 & 0 & 0 
    \end{pmatrix} \TR{y}(\theta) = \big[ \evec{R}_y \big] \TR{y}(\theta), \label{eq:ERy} \\
    \dfrac{\mathrm{d} \TR{z}(\theta)}
          {\mathrm{d} \theta}
    &=     
    \begin{pmatrix}
        0 & -1 & 0 & 0 \\
        1 & 0 & 0 & 0 \\
        0 & 0 & 0 & 0 \\
        0 & 0 & 0 & 0  
    \end{pmatrix} \TR{z}(\theta) = \big[ \evec{R}_z \big] \TR{z}(\theta), \label{eq:ERz}
\end{align}
for a revolute joint, or one of
\begin{align}
    \dfrac{\mathrm{d} \Tt{x}(d)}
          {\mathrm{d} d}
    &=     
    \begin{pmatrix}
        0 & 0 & 0 & 1 \\
        0 & 0 & 0 & 0 \\
        0 & 0 & 0 & 0 \\
        0 & 0 & 0 & 0 
    \end{pmatrix} = \sk{\evec{t}_x}, \label{eq:ETx}\\
    \dfrac{\mathrm{d} \Tt{y}(d)}
          {\mathrm{d} d}
    &=     
    \begin{pmatrix}
        0 & 0 & 0 & 0 \\
        0 & 0 & 0 & 1 \\
        0 & 0 & 0 & 0 \\
        0 & 0 & 0 & 0 
    \end{pmatrix} = \sk{\evec{t}_y}, \label{eq:ETy}\\
    \dfrac{\mathrm{d} \Tt{z}(d)}
          {\mathrm{d} d}
    &=     
    \begin{pmatrix}
        0 & 0 & 0 & 0 \\
        0 & 0 & 0 & 0 \\
        0 & 0 & 0 & 1 \\
        0 & 0 & 0 & 0 
    \end{pmatrix} = \sk{\evec{t}_z}, \label{eq:ETz}
\end{align}
for a prismatic joint. For the second derivative of an elementary transform with respect to the same joint variable, as is the case for $k = j$ in (\ref{eq:ddets}), the result is
\begin{align}
    \dfrac{\mathrm{d}^2 \TR{x}(\theta)}
          {\mathrm{d} \theta^2}
    &=     
    \big[ \evec{R}_x \big]^2 \ \TR{x}(\theta), \label{eq:ddERx} \\
    \dfrac{\mathrm{d}^2 \TR{y}(\theta)}
          {\mathrm{d} \theta^2}
    &=     
    \big[ \evec{R}_y \big]^2 \  \TR{y}(\theta), \label{eq:ddERy} \\
    \dfrac{\mathrm{d}^2 \TR{z}(\theta)}
          {\mathrm{d} \theta^2}
    &=     
    \big[ \evec{R}_z \big]^2 \  \TR{z}(\theta), \label{eq:ddERz}
\end{align}
for a revolute joint or a zero matrix for a prismatic joint.

As for the manipulator Jacobian in Part 1, to form the manipulator Hessian, we partition it into translational and rotational components as expressed in (\ref{eq:a1}).

To form $\mat{H}_{\alpha_{jk}}$, the angular component of the manipulator Hessian of a joint variable $j$ with respect to another joint variable $k$, we take the partial derivative of the $j^{th}$ column of the manipulator Jacobian in 
\begin{align} \label{eq:jwj1}
    \Jwx{j}(\q) 
    &=
    \iskx{
        \rho
        \left(
            \frac{\partial \matfn{T}{\vec{q}}}
                {\partial q_j} 
        \right)
        \rho
        \left(
            \matfn{T}{\vec{q}}
        \right)^\top
    }
\end{align}
using the product rule
\begin{align} \label{eq:hess0}
    \Hwx{jk}
    &=
    \dfrac{\partial \Jwx{j}(\q)}
        {\partial q_k} \nonumber \\
    &=
    \vee_\times \bigg(
        \rho
        \left(
            \frac{\partial^2 \matfn{T}{\vec{q}}}
                {\partial q_j \partial q_k} 
        \right)
        \rho
        \left(
            \matfn{T}{\vec{q}}
        \right)^\top
    \bigg.
    + \nonumber \\
    & \qquad \quad
    \bigg.
        \rho
        \left(
            \frac{\partial \matfn{T}{\vec{q}}}
                {\partial q_j} 
        \right)
        \rho
        \left(
            \frac{\partial \matfn{T}{\vec{q}}}
                {\partial q_k} 
        \right)^\top
    \bigg) \nonumber \\
    &=
    \iskx{
        \HR_{jk} \
        \rho
        \left(
            \matfn{T}{\vec{q}}
        \right)^\top
        +
        \JR_j
        \JR_k^\top
    }
\end{align}
where $\Hwx{jk} \in \mathbb{R}^3$, and $\HR_{jk}$ is obtained from (\ref{eq:ddets}).

To form $\mat{H}_{a_{jk}}$, the translational component of the manipulator Hessian for joint variable $j$ with respect to another joint variable $k$, we take the partial derivative of the $j^{th}$ translational component of the manipulator Jacobian in
\begin{align} \label{eq:jvj1}
    \Jvx{j}(\q) 
    &=
    \tau
    \left(
        \frac{\partial \matfn{T}{\vec{q}}}
            {\partial q_j} 
    \right)
\end{align}
which provides
\begin{align} \label{eq:hess1}
    \Hvx{jk}
    &=
    \dfrac{\partial \Jv{_{j}}(\q)}
        {\partial q_k} \nonumber \\
    &=
    \tau
    \left(
        \frac{\partial^2 \matfn{T}{\vec{q}}}
            {\partial q_j \partial q_k} 
    \right) \nonumber \\
    &= 
    \Ht_{jk}
\end{align}
where $\H_{a_{jk}} \in \mathbb{R}^3$, and $\Ht_{jk}$ is obtained from (\ref{eq:ddets}).

Stacking (\ref{eq:hess0}) and (\ref{eq:hess1}) we form the component of the manipulator Hessian for joint variable $j$ with respect to another joint variable $k$

\begin{align} \label{eq:hess2}
    \H_{jk}
    &=
    \begin{pmatrix}
        \Hvx{jk} \\
        \Hwx{jk}
    \end{pmatrix}
\end{align}
where $\H_{jk} \in \mathbb{R}^{6}$.

The component of the manipulator Hessian for joint variable $j$ is formed by arranging (\ref{eq:hess2}) into columns of a matrix

\begin{align} \label{eq:hess3}
    \H_j
    &=
    \begin{pmatrix}
        \H_{j1} &
        \cdots &
        \H_{jn}
    \end{pmatrix}
\end{align}
where $\H_{j} \in \mathbb{R}^{6 \times n}$.

The whole manipulator Hessian is formed by arranging (\ref{eq:hess3}) into \textit{slices} of a tensor

\begin{align} \label{eq:hess4}
    \H
    &=
    \begin{pmatrix}
        \H_1 &
        \cdots &
        \H_n
    \end{pmatrix}
\end{align}
where $\H \in \mathbb{R}^{n \times 6 \times n}$ and last last two dimensions of $\H$ define the dimension of the slices $\H_i$. We show the formation of the manipulator Hessian for a 7-joint manipulator in Figure \ref{fig:hessian}.

\subsection*{Fast Manipulator Hessian} \label{sec:ddHf}

We can calculate the manipulator Hessian using (\ref{eq:hess0}) and (\ref{eq:hess1}) with (\ref{eq:ddets}), however this has $\mathcal{O}(n^3)$ time complexity.

We revisit (\ref{eq:hess0}) while substituting in
\begin{align} \label{eq:dr1}
    \frac{\mathrm{d}\mat{R}(\theta)}
        {\mathrm{d} \theta} 
    &= \skx{\evec{\omega}} \mat{R}(\theta(t)) 
\end{align}
and simplify
\begin{align}
    \Hwx{jk}
    &=
    \vee_\times \bigg(
        {\skx{\evec{\omega}_k}} {\skx{\evec{\omega}_j}} \R(\q)
        \matfn{R}{\q}^\top + 
    \bigg. \nonumber \\
    & \qquad \
    \bigg.
        {\skx{\evec{\omega}_j}} \R(\q)
        \left(
        {\skx{\evec{\omega}_k}} \R(\q)
        \right)^\top
    \bigg) \nonumber \\
    &=
    \iskx{
        {\skx{\evec{\omega}_k}} {\skx{\evec{\omega}_j}} + 
        {\skx{\evec{\omega}_j}} \R(\q) \R(\q)^\top
        {\skx{\evec{\omega}_k}}^\top
    } \nonumber \\
    &=
    \iskx{
        {\skx{\evec{\omega}_k}} {\skx{\evec{\omega}_j}} - 
        {\skx{\evec{\omega}_j}}
        {\skx{\evec{\omega}_k}}
    }.
\end{align}

Since we know that $\Jw_x = {\skx{\evec{\omega}_x}}$, and using the identity $\skx{\vec{a}\times \vec{b}} = \skx{\vec{a}}\skx{\vec{b}}-\skx{\vec{b}}\skx{\vec{a}}$ we show that

\begin{align}
    \Hwx{jk}
    &= \iskx{\skx{\Jwx{k}} \skx{\Jwx{j}} - \skx{\Jwx{j}} \skx{\Jwx{k}}} \nonumber \\
    &= \Jwx{k} \times \Jwx{j}
\end{align}
which means that the rotational component of the manipulator Hessian can be calculated from the rotational components of the manipulator Jacobian. 
A key relationship is that the velocity of joint $j$, with respect to the velocity of the same, or preceding joint $k$, does not contribute acceleration to the end-effector from the perspective of joint $j$. Consequently, $\Hwx{jk}=0$ when $k\geq j$.

For the translational component of the manipulator Hessian $\Hv$, we can see in (\ref{eq:ddets}) that two of the conditions will have the same result: when $k < j$, and when $k > j$. Therefore, we have

\begin{align}
    \Hvx{jk}(\q) = \Hvx{kj}(\q)
\end{align}
and by exploiting this relationship, we can simplify (\ref{eq:hess1}) to

\begin{align}
    \Hvx{jk}(\q)
    &= \skx{\Jwx{k}} \Jvx{j} \nonumber \\
    &= \Jwx{a} \times \Jvx{b}
\end{align}
where $a = \min(j,k)$, and $b = \max(j,k)$. This means that the translational component of the manipulator Hessian can be calculated from components of the manipulator Jacobian.

Through this simplification, computation of the manipulator Hessian reduces to $\mathcal{O}(n^2)$ time complexity.


\section*{Deriving Higher Order Derivatives}

Obtaining the $n^{th}$ partial derivative of the manipulator kinematics, where $n \geq 3$, can be obtained using the product rule on the $(n - 1)^{th}$ partial derivative while considering their partitioned form.

For example, to obtain the $3^{rd}$ partial derivative, we take the partial derivative of the manipulator Hessian with respect to the joint coordinates, in its partitioned form

\begin{align}
    \frac{\partial \mat{H}_{jk}(\vec{q})}
         {\partial q_l}
    &=
    \begin{pmatrix}
        \dfrac{\partial \mat{H}_{a_{jk}}(\vec{q})}
              {\partial q_l} \\
        \dfrac{\partial \mat{H}_{\alpha_{jk}}(\vec{q})}
              {\partial q_l}
    \end{pmatrix} \nonumber \\
    &=
    \begin{pmatrix}
        \dfrac{\partial }
              {\partial q_l}
        \left( \Jwx{k} \times \Jwx{j} \right) \\
        \dfrac{\partial}
              {\partial q_l}
        \left( \Jwx{k} \times \Jvx{j} \right)
    \end{pmatrix} \nonumber \\
    &=
    \begin{pmatrix}
        \left( \Hwx{kl} \times \Jwx{j} \right) +
        \left( \Jwx{k} \times \Hwx{jl} \right) \\
        \left( \Hwx{kl} \times \Jvx{j} \right) +
        \left( \Jwx{k} \times \Hvx{jl} \right)
    \end{pmatrix}
\end{align}
where $\frac{\partial \mat{H}_{jk}(\vec{q})}{\partial q_l} \in \mathbb{R}^{6}$. Continuing, we obtain the following

\begin{align}
    \frac{\partial \mat{H}_{jk}(\vec{q})}
         {\partial \vec{q}}
    &=
    \begin{pmatrix}
        \dfrac{\partial \mat{H}_{jk}(\vec{q})}
            {\partial q_0} & \hdots & 
        \dfrac{\partial \mat{H}_{jk}(\vec{q})}
            {\partial q_n}
    \end{pmatrix}
\end{align}
where $\frac{\partial \mat{H}_{jk}(\vec{q})}{\partial q_l} \in \mathbb{R}^{6 \times n}$,

\begin{align}
    \frac{\partial \mat{H}_{j}(\vec{q})}
        {\partial \vec{q}}
    &=
    \begin{pmatrix}
        \dfrac{\partial \mat{H}_{j_0}(\vec{q})}
            {\partial \vec{q}} & \hdots & 
        \dfrac{\partial \mat{H}_{j_n}(\vec{q})}
            {\partial \vec{q}}
    \end{pmatrix}
\end{align}
where $\frac{\partial \mat{H}_{jk}(\vec{q})}{\partial \vec{q}} \in \mathbb{R}^{n \times 6 \times n}$, and finally 

\begin{align}
    \frac{\partial \mat{H}(\vec{q})}
         {\partial \vec{q}}
    &=
    \begin{pmatrix}
        \dfrac{\partial \mat{H}_{0}(\vec{q})}
            {\partial \vec{q}} & \hdots & 
        \dfrac{\partial \mat{H}_{n}(\vec{q})}
            {\partial \vec{q}}
    \end{pmatrix}
\end{align}
where $\frac{\partial \mat{H}(\vec{q})}{\partial \vec{q}} \in \mathbb{R}^{n \times n \times 6 \times n}$ is the 4-dimensional tensor representing the $3^{rd}$ partial derivative of the manipulator kinematics.

We have included a function as part of our open source Robotics Toolbox for Python \cite{rtb} which can calculate the $n^{th}$ partial derivative of the manipulator kinematics. Note that the function has $\mathcal{O}(n^{order})$ time complexity, where $order$ represents the order of the partial derivative being calculated.


\section*{Analytical Form}

The kinematic derivatives we have presented so far have been in geometric form with translational velocity $\vec{v}$ and angular velocity $\vec{\omega}$ vectors and their derivatives. Some applications require the manipulator Jacobian and further derivatives to be expressed with different orientation rate representations such as the rate of change of roll-pitch-yaw angles, Euler angles or exponential coordinates -- these are called analytical forms.

One important application that requires this is task space dynamics \cite{peter} and operational space control \cite{osc}. Operational space control is a dynamics formulation for tasks that require constrained end-effector motion and force control. There are many everyday tasks that can make use of this control formulation such as opening a door or cleaning a surface. While dynamics are outside the scope of this tutorial, these dynamics controllers require the manipulator Jacobian and further derivatives to be represented in analytical form.

\subsection*{Roll-Pitch-Yaw Analytical Form}

For XYZ roll-pitch-yaw angles in $\vec{\varGamma} = (\alpha, \beta, \gamma)$, the resulting rotation matrix is
\begin{align} \label{eq:rpy}
    \mat{R}
    &= \mat{R}_x(\gamma) \mat{R}_y(\beta) \mat{R}_z(\alpha) \\
    &=
    \begin{pmatrix}
        c\beta c\alpha & -c\beta s\alpha &   c\beta \\
        c\gamma s\alpha + c\alpha s\beta s\gamma & -s\beta s\gamma s\alpha + c\gamma c\alpha & -c\beta s\gamma \\
        s\gamma s\alpha - c\gamma c\alpha s\beta & c\gamma s\beta s\alpha + c\alpha s\gamma  & c\beta c\gamma
    \end{pmatrix} \nonumber
\end{align}
where $s\theta$ and $c\theta$ are short for $\sin(\theta)$ and $\cos(\theta)$. From Part 1 we have the relationship
\begin{align}
    \dmat{R} &= \skx{\vec{\omega}} \mat{R} \nonumber \\
    \dmat{R} \mat{R}^\top &= \skx{\vec{\omega}} \nonumber \\
    \iskx{\dmat{R} \mat{R}^\top} &= \vec{\omega}
\end{align}
which gives us the result
\begin{align}
    \vec{\omega} =
    \begin{pmatrix}
        \omega_x \\
        \omega_y \\
        \omega_z \\
    \end{pmatrix}
    &=
    \begin{pmatrix}
        s\beta \dot{\alpha} + \dot{\gamma} \\
        -c\beta s\gamma \dot{\alpha} + c\gamma \dot{\beta} \\
        c\beta c\gamma \dot{\alpha} + s\gamma \dot{\beta}
    \end{pmatrix} \nonumber \\
    &=
    \begin{pmatrix}
        s\beta & 0 & 1 \\
        -c\beta s\gamma & c\gamma & 0 \\
        c\beta c\gamma & s\gamma & 0
    \end{pmatrix}
    \begin{pmatrix}
        \dot{\alpha} \\
        \dot{\beta} \\
        \dot{\gamma}
    \end{pmatrix} \nonumber \\
    &= \mat{A}(\vec{\varGamma})\dvec{\varGamma} \label{eq:jrpy}
\end{align}
where $\mat{A}(\vec{\varGamma})$ is a Jacobian that maps XYZ roll-pitch-yaw angle rates to angular velocity.

The analytical Jacobian represented with roll-pitch-yaw angle rates is
\begin{align} \label{eq:ja}
    \J_\Lambda(\q)(\vec{q})
    &=
    \mat{J}_\Gamma(\vec{\varGamma}) \ \J(\q) \nonumber 
    \\
    &=
    \begin{pmatrix}
        \vec{1}_{3 \times 3} & \vec{0}_{3 \times 3} \\
        \vec{0}_{3 \times 3} & \mat{A}^{-1}(\vec{\varGamma}) \\
    \end{pmatrix}
    \mat{J}(\vec{q}).
\end{align}

In the case where $\beta = \pm 90\deg$, $\mat{A}$ will be singular and its inverse does not exist. Therefore, for applications involving the analytical differential kinematics, it is important to choose an angular representation where the singularity lies outside of the normal operating range of the robot \cite{peter}.

The derivative of $\J_\Lambda(\q)$ is typically used in applications that have a task-space acceleration term. We can obtain $\dmat{J}_\Lambda(\q)$ from (\ref{eq:ja}) using the product rule

\begin{align} \label{eq:dja}
    \dmat{J}_\Lambda(\vec{q})
    &=
    \dfrac{\mathrm{d} \mat{J}_\Gamma(\vec{\varGamma})}
          {\mathrm{d} t} \
    \J(\q) +
    \mat{J}_\Gamma(\vec{\varGamma}) \ 
    \dfrac{\mathrm{d} \J(\q)}
          {\mathrm{d} t} \nonumber \\
    &=
    \dfrac{\mathrm{d} \mat{J}_\Gamma(\vec{\varGamma})}
          {\mathrm{d} t} \
    \J(\q) +
    \mat{J}_\Gamma(\vec{\varGamma}) \ 
    \big(
        \H(\q) \dvec{q}
    \big)
\end{align}
where the derivative of the augmented Jacobian $\mat{J}_\Gamma(\vec{\varGamma})$ is
\begin{align} \label{eq:ja2}
    \dfrac{\mathrm{d} \mat{J}_\Gamma(\vec{\varGamma})}
          {\mathrm{d} t}
    &=
    \begin{pmatrix}
        \vec{0}_{3 \times 3} & \vec{0}_{3 \times 3} \\
        \vec{0}_{3 \times 3} & \dmat{A}^{-1}(\vec{\varGamma}, \ \dvec{\varGamma}) \\
    \end{pmatrix}.
\end{align}

As previously mentioned, the analytical Jacobian can be derived for different orientation parameterisations including Euler angles, exponential coordinates or ZYX roll-pitch-yaw angles.
To achieve this, the rotation matrix in (\ref{eq:rpy}) is replaced with the appropriate elements for the different parameterisation and following through the methodology presented in this section to produce the appropriate analytical Jacobian and derivative.


\section*{Advanced Velocity Control}

Many modern manipulators are redundant -- they have more than six degrees of freedom. We can exploit this redundancy by having the robot optimise some performance measure while still achieving the original goal. In this section, we start with the resolved-rate motion control (RRMC) algorithm explained in Part 1 of this tutorial
\begin{equation} \label{eq:rrmc}
     \dvec{q} = \matfn{J}{\vec{q}}^{+} \ \vec{\nu}.
\end{equation}

\subsection*{Null-space Projection}

The Jacobian of a redundant manipulator has a null space. Any joint velocity vector which is a linear combination of the manipulator Jacobian's null-space basis vectors will result in zero end-effector motion ($\vec{\nu} = 0$). We can augment (\ref{eq:rrmc}) to add a joint velocity vector $\dvec{q}_{null}$ which can be projected into the null space resulting in zero end-effector spatial velocity
\begin{equation} \label{eq:ns0}
    \dvec{q} =
    \underbrace{
        \matfn{J}{\vec{q}}^{+} \ \vec{\nu}
    }_{\mathrm{end-effector \ motion}} + \ \
        \underbrace{
        \left(
            \mat{1}_n - \mat{J}(\vec{q})^{+} \mat{J}(\vec{q})
        \right)
        \dvec{q}_\mathrm{null}
    }_{\mathrm{null-space \ motion}}
\end{equation}
where $\dvec{q}_\mathrm{null}$ is the desired joint velocities for the null-space motion.

We can set $\dvec{q}_\mathrm{null}$ to be the gradient of any scalar performance measure $\gamma(\vec{q})$ where the performance measure is a differentiable function of the joint coordinates $\vec{q}$.

Park \cite{gpm} proposed using the gradient of the Yoshikawa manipulability index as $\dvec{q}_\mathrm{null}$ in (\ref{eq:ns0}). As detailed in Part 1 of this tutorial, the manipulability index \cite{manip} is calculated as
\begin{align} \label{eq:man0}
    m(\vec{q}) = \sqrt{
        \mbox{det}
        \Big(
            \emat{J}(\vec{q})
            \emat{J}(\vec{q})^\top
        \Big)
    }
\end{align}
where $\emat{J}(\vec{q}) \in \mathbb{R}^{3 \times n}$ is either the translational or rotational rows of $\mat{J}(\vec{q})$ causing  $m(\vec{q})$ to describe the corresponding component of manipulability. Excurse \ref{fig:note1} highlights considerations for manipulators with mixed joint types.

\begin{excurse}[!b]

    \noindent\textcolor{lineblue}{\rule{\linewidth}{2pt}}
    \vspace{-13pt}
    
    \noindent\colorbox{boxblue}{
        
        \parbox{\linewidth-12.5pt}{
            \vspace{4pt}
            \caption{
                \usefont{OT1}{ugq}{n}{n}{Mixed Joint Manipulators}
            }
            \label{fig:note1}
            \vspace{-5pt}

            \color{dgrey}

            Manipulators that contain different joint types -- such as those that contain both revolute and prismatic joints -- have implications on certain algorithms. Scaling issues can be introduced due to the different units which may cause a translation or rotation component to dominate the result. In the case of using the manipulator Jacobian for a performance metric, care must be taken to ensure the performance is acceptable or use a scaling approach \cite{stocco1999use}. In the case where a gain is applied to a control input on each joint, simply use an appropriately scaled gain value for each different joint type.
    
            \vspace{3pt}
        }
    }
    
    \vspace{-1pt}
    \noindent\textcolor{lineblue}{\rule{\linewidth}{2pt}} 
\end{excurse}

Taking the time derivative of (\ref{eq:man0}), using the chain rule
\begin{align} \label{eq:man1}
    \frac{\mathrm{d} \ m(t)}
            {\mathrm{d} t} = 
    \dfrac{1}
            {2m(t)} 
    \frac{\mathrm{d}}
         {\mathrm{d} t} 
    \mbox{det} \left( \emat{J}(\vec{q}) \emat{J}(\vec{q})^\top \right)
\end{align}
we can write this as
\begin{align} \label{eq:man2}
    \dot{m}
    &=
    \vec{J}_m^\top(\vec{q}) \ \dvec{q}
\end{align}
where
\begin{equation} \label{eq:man3}
    \vec{J}_m^\top(\vec{q})
    =
    m
    \begin{pmatrix}
        \mbox{vec} \left( \emat{J}(\vec{q}) \emat{H}_1(\vec{q})^\top \right)^\top 
        \mbox{vec} \left( (\emat{J}(\vec{q})\emat{J}(\vec{q})^\top)^{-1} \right) \\
        \mbox{vec} \left( \emat{J}(\vec{q}) \emat{H}_2(\vec{q})^\top \right)^\top 
        \mbox{vec} \left( (\emat{J}(\vec{q})\emat{J}(\vec{q})^\top)^{-1} \right) \\
        \vdots \\
        \mbox{vec} \left( \emat{J}(\vec{q}) \emat{H}_n(\vec{q})^\top \right)^\top 
        \mbox{vec} \left( (\emat{J}(\vec{q})\emat{J}(\vec{q})^\top)^{-1} \right) \\
    \end{pmatrix}
\end{equation}
is the manipulability Jacobian $\vec{J}^\top_m \in \mathbb{R}^n$ and where the vector operation $\mbox{vec}(\cdot) : \mathbb{R}^{a \times b} \rightarrow \mathbb{R}^{ab}$ converts a matrix column-wise into a column vector,
and $\emat{H}_i \in \mathbb{R}^{3 \times n}$ is the translational or rotational component (matching the choice of $\emat{J}(\vec{q})$)
of $\mat{H}_i \in \mathbb{R}^{6 \times n}$ which is the $i^{th}$ component of the manipulator Hessian tensor $\mat{H} \in \mathbb{R}^{n \times 6 \times n}$.

The complete equation proposed by Park \cite{gpm} is
\begin{equation} \label{eq:ns1}
    \dvec{q} =
    \matfn{J}{\vec{q}}^{+} \ \vec{\nu} +
    \frac{1}{\lambda} 
    \Big(
        \left(
            \mat{1}_n - \mat{J}(\vec{q})^{+} \mat{J}(\vec{q})
        \right)
        \mat{J_m}(\vec{q})
    \Big)
\end{equation}
where $\lambda$ is a gain that scales the magnitude of the null-space velocities. This equation will choose joint velocities $\dvec{q}$ which will achieve the end-effector spatial velocity $\vec{\nu}$ while also improving the translational and/or rotational manipulability of the robot.

\begin{figure}[!t]
    \centering
    \includegraphics[height=4.95cm]{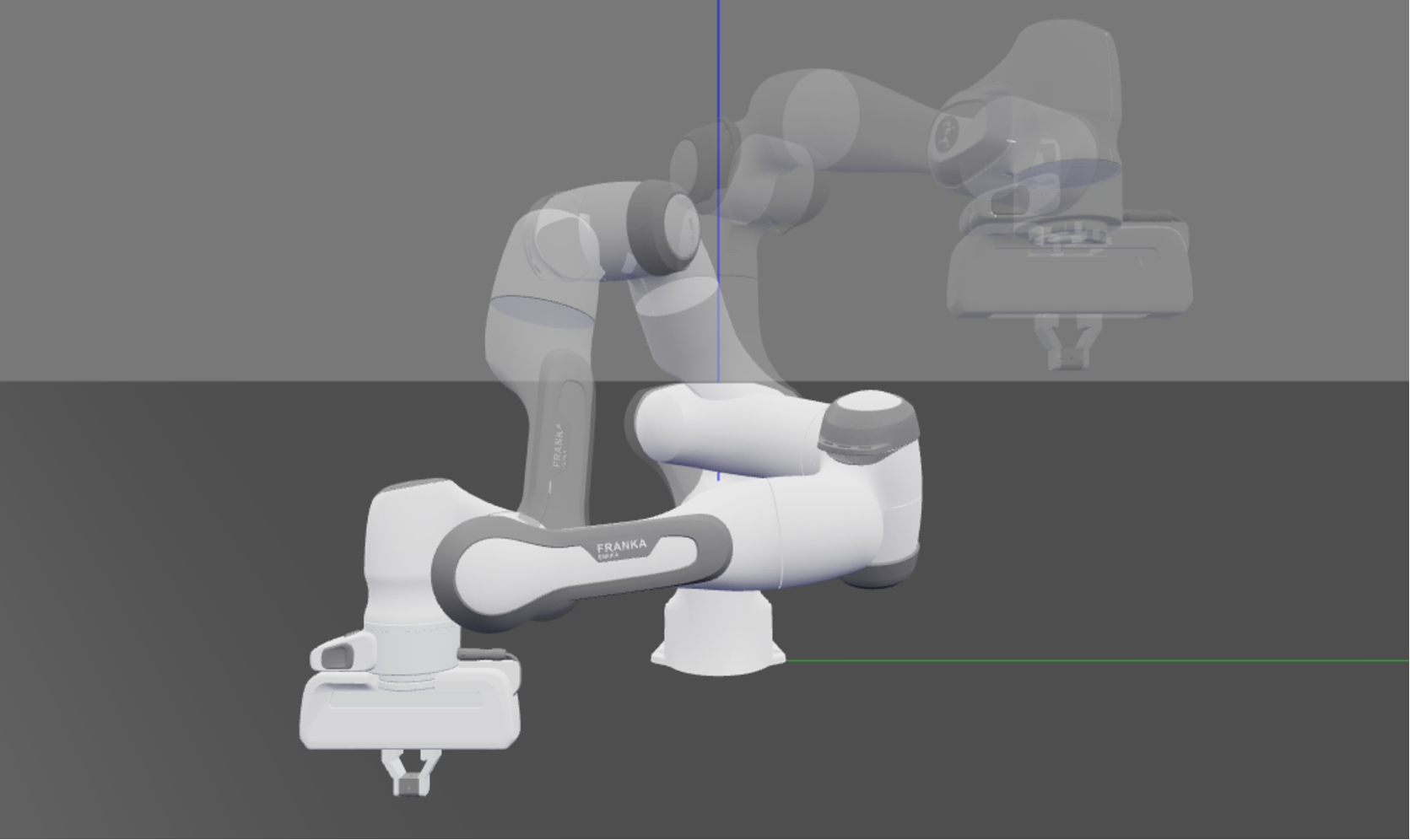}
    \caption{
        Comparison of a Panda robot which has been controlled by RRMC (semi-transparent robot) and the manipulability maximising controller in (\ref{eq:ns1}) (opaque robot). The transparent robot displays the initial pose.
    }
    \label{fig:gpm1}
\end{figure}

\begin{figure}[!t]
    \centering
    \includegraphics[height=4.95cm]{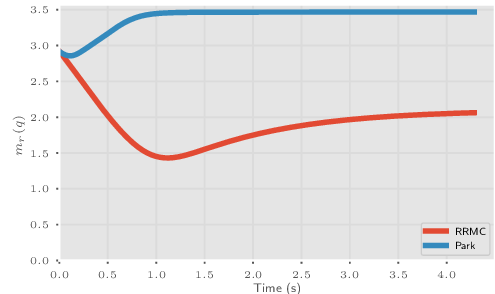}
    \caption{
        The rotational manipulability of the robot being controlled by the RRMC and Park controllers.
    } 
    \label{fig:gpm2}
\end{figure}

As with RRMC, (\ref{eq:ns1}) will provide the joint velocities for a desired end-effector velocity. As we did in Part 1, we can employ (\ref{eq:ns1}) in a position-based servoing (PBS) controller to drive the end-effector towards some desired pose. The PBS scheme is
\begin{align} \label{eq:pbs2}
    \vec{\nu} = \vec{k} \vec{e}
\end{align}
where $\vec{\nu}$ is the desired end-effector spatial velocity to be used in (\ref{eq:ns1}), and the choice and calculation of $\vec{k}$ and $\vec{e}$ is detailed in (37)-(43) of Part 1. Figure \ref{fig:gpm1} shows the difference in final joint configuration between RRMC and Park's controller (using the rotational manipulability Jacobian) where both controllers are employed with a PBS controller to define the demanded end-effector velocity to achieve the same goal end-effector pose. Figure \ref{fig:gpm2} compares the rotational manipulability of the controllers throughout the trajectory. We can see that, as opposed to the RRMC controller, the Park controller increases the initial rotational manipulability and maintains the higher value throughout the trajectory.

Null-space projection is not limited to manipulability maximisation. Any subtask which can be expressed as a differentiable function of $\vec{q}$ can be used, and multiple sub-tasks can be individually weighted and added together to be used as $\dvec{q}_\mathrm{null}$. For example, Baur \cite{gpm2} used the manipulability Jacobian and an additional weighting on joint positions which discouraged joints from getting too close to their physical limits.

\subsection*{Quadratic Programming}

In this section, we are going to redefine our motion controllers as a quadratic programming (QP) optimisation problem rather than a matrix equation. In general, a constrained QP is formulated as \cite{opt0}
\begin{align} \label{eq:qp0}
    \min_x \quad f_o(\vec{x}) &= \frac{1}{2} \vec{x}^\top \mat{Q} \vec{x}+ \vec{c}^\top \vec{x}, \\ 
    \mbox{subject to} \quad \mat{A}_1 \vec{x} &= \vec{b}_1, \nonumber \\
    \mat{A}_2 \vec{x} &\leq \vec{b}_2, \nonumber \\
    \vec{g} &\leq  \vec{x} \leq \vec{h}. \nonumber 
\end{align}
where $f_o(\vec{x})$ is the objective function which is subject to the equality and inequality constraints, and $\vec{h}$ and $\vec{g}$ represent the upper and lower bounds of $\vec{x}$. A quadratic program is strictly convex when the matrix $\mat{Q}$ is positive definite \cite{opt0}. This framework allows us to solve the same problems as above, but with additional flexibility. In practice, the downside to this QP approach is that it is marginally more complex to implement.

We can rewrite RRMC (\ref{eq:rrmc}) as a QP
\begin{align} \label{eq:qp1}
	\min_{\dvec{q}} \quad f_o(\dvec{q}) 
	&= 
    \frac{1}{2} 
    \dvec{q}^\top 
    \mat{1}_n 
    \dvec{q}, \\
	\mbox{subject to} \quad
	\matfn{J}{\vec{q}} \dvec{q} &= \vec{\nu},
	\nonumber \\
	\dvec{q}^- &\leq \dvec{q} \leq \dvec{q}^+ \nonumber 
\end{align}
where we have imposed $\dvec{q}^-$ and $\dvec{q}^+$ as upper and lower joint-velocity limits. If the manipulator has more degrees of freedom than necessary to reach its entire task space, the QP will achieve the desired end-effector velocity with the minimum joint-velocity norm (the same result as the pseudoinverse solution, (36) in Part 1). If the manipulator has six joints, then the solution will be the same as (35) in Part 1.

We can rewrite Park's controller as
\begin{align} \label{eq:qp3}
    \min_{\dvec{q}} \quad f_o(\dvec{q}) 
    &= 
    \frac{1}
        {2} 
    \dvec{q}^\top \lambda \mat{1}_n \dvec{q} - \vec{J}_m^\top(\vec{q}) \dvec{q}, \\ 
    \mbox{subject to} \quad \matfn{J}{\vec{q}} \dvec{q} &= \vec{\nu}. \nonumber
\end{align}
where the manipulability Jacobian fits into the linear component of the objective function.

This was extended in \cite{mmc} with velocity dampers to enable joint-position-limit avoidance. Velocity dampers \cite{pp0} are used to constrain velocities and dampen an input velocity as some position limit is approached. The velocity is only damped if it is within some influence distance of the limit. A joint velocity is constrained to prevent joint limits using a velocity damper constraint
\begin{align} \label{eq:qp4}
    \dot{q} \leq
    \eta
    \frac{\rho - \rho_s}
        {\rho_i - \rho_s}
    \qquad \mbox{if} \ \rho < \rho_i
\end{align}
where $\rho \in \mathbb{R}^+$ is the distance or angle to the nearest joint limit, $\eta \in \mathbb{R}^+$ is a gain that adjusts the aggressiveness of the damper, $\rho_i$ is the influence distance within which to activate the damper, and $\rho_s$ is the stopping distance in which the distance $\rho$ will never be able to reach or enter.
In a robot with mixed joint types see Excurse \ref{fig:note1}.
We can stack velocity dampers to perform joint position limit avoidance for each joint within a robot, and incorporate it into our QP (\ref{eq:qp3}) as an inequality constraint
\begin{align} \label{eq:qp5}
    \mat{1}_n
    \dvec{q} &\leq
    \eta
    \begin{pmatrix}
        \frac{\rho_0 - \rho_s}
            {\rho_i - \rho_s} \\
        \vdots \\
        \frac{\rho_n - \rho_s}
            {\rho_i - \rho_s} 
    \end{pmatrix}
\end{align}
where the identity $\mat{1}_n$ is included to show how the equation fits into the general form $\mat{A}\vec{x} \leq \vec{b}$ of an inequality constraint.

It is possible that the robot will fail to reach the goal when the constraints create local minima. In such a scenario, the error $\vec{e}$ in the PBS scheme is no longer decreasing and the robot can no longer progress due to the constraints in (\ref{eq:qp5}).

The methods shown up to this point require a redundant robot, where the number of joints is greater than 6. In \cite{mmc}, extra redundancy was introduced to the QP by relaxing the equality constraint in (\ref{eq:qp5}) to allow for intentional deviation, or \emph{slack} from the desired end-effector velocity $\vec{\nu}$. The slack has the additional benefit of giving the solver extra flexibility to meet constraints and avoid local minima. The augmented QP is defined as
\begin{align} \label{eq:mmc}
    \min_x \quad f_o(\vec{x}) &= \frac{1}{2} \vec{x}^\top \mathcal{Q} \vec{x}+ \mathcal{C}^\top \vec{x}, \\ 
    \mbox{subject to} \quad \mathcal{J} \vec{x} &= \vec{\nu}, \nonumber \\
    \mathcal{A} \vec{x} &\leq \mathcal{B}, \nonumber \\
    \vec{x}^- &\leq \vec{x} \leq \vec{x}^+ \nonumber
\end{align}
with
\begin{align}
    \vec{x} &= 
    \begin{pmatrix}
        \dvec{q} \\ \vec{\delta}
    \end{pmatrix} \in \mathbb{R}^{(n+6)}  \\
    \mathcal{Q} &=
    \begin{pmatrix}
        \lambda_q \mat{1}_{n} & \mathbf{0}_{6 \times 6} \\ \mathbf{0}_{n \times n} & \lambda_\delta \mat{1}_{6}
    \end{pmatrix} \in \mathbb{R}^{(n+6) \times (n+6)} \\
    \mathcal{J} &=
    \begin{pmatrix}
        \mat{J}(\vec{q}) & \mat{1}_{6}
    \end{pmatrix} \in \mathbb{R}^{6 \times (n+6)} \label{eq:gain} \\
    \mathcal{C} &= \label{eq:jm}
    \begin{pmatrix}
        \vec{J}_m \\ \mat{0}_{6 \times 1}
    \end{pmatrix} \in \mathbb{R}^{(n + 6)} \\
    \mathcal{A} &= 
    \begin{pmatrix}
        \mat{1}_{n \times n + 6} \\
    \end{pmatrix} \in \mathbb{R}^{(l + n) \times (n + 6)} \\
    \mathcal{B} &= \label{eq:gain2}
    \eta
    \begin{pmatrix}
        \frac{\rho_0 - \rho_s}
                {\rho_i - \rho_s} \\
        \vdots \\
        \frac{\rho_n - \rho_s}
                {\rho_i - \rho_s} 
    \end{pmatrix} \in \mathbb{R}^{n} \\
    \vec{x}^{-, +} &= 
    \begin{pmatrix}
        \dvec{q}^{-, +} \\
        \vec{\delta}^{-, +}
    \end{pmatrix} \in \mathbb{R}^{(n+6)},
\end{align}
where
$\vec{\delta} \in \mathbb{R}^6$ is the slack vector,
$\lambda_\delta \in \mathbb{R}^+$ is a gain term that adjusts the cost of the norm of the slack vector in the optimiser,
$\dvec{q}^{-,+}$ are the minimum and maximum joint velocities, and 
$\vec{\delta}^{-,+}$ are the minimum and maximum slack velocities.
Each of the gains can be adjusted dynamically. For example, in practice $\lambda_\delta$ is typically large when far from the goal, but reduces towards 0 as the goal approaches.

The effect of this augmented optimisation problem is that the equality constraint is equivalent to
\begin{equation} \label{eq:qp2}
    \vec{\nu}(t) - \vec{\delta}(t) = \matfn{J}{\vec{q}} \dvec{q}(t)
\end{equation}
which clearly demonstrates that the slack is essentially intentional error, where the optimiser can choose to move components of the desired end-effector motion into the slack vector.
For both redundant and non-redundant robots, this means that the robot may stray from the straight-line motion to improve manipulability and avoid a singularity, avoid running into joint position limits or stay bound by the joint velocity limits.

Velocity dampers are further demonstrated in \cite{neo} where they are used to incorporate real-time obstacle avoidance into the QP. Furthermore, in \cite{holistic} the QP framework was extended to allow for holistic differential-kinematic control of a mobile manipulator.


\section*{Advanced Inverse Kinematics}

In Part 1 of this tutorial, we considered unconstrained numerical inverse kinematics. In this section, we are going to extend this to consider constraints such as joint position limits and introduce some performance measures.

To begin, we will first consider the inverse kinematics solver based on the Newton-Raphson (NR) method which we described in Part 1 as the iteration
\begin{align} \label{eq:ik1}
    \vec{q}_{k+1} = \vec{q}_k + {\mat{J}(\vec{q}_k)}^{+} \vec{e}_k
\end{align}
until the desired end-effector pose is reached, where $\vec{e}$ is the position and angle-axis error between the current pose and the desired end-effector pose ((37) of Part 1) expressed in the world frame, and $\mat{J} = {^0\mat{J}}$ is the base-frame manipulator Jacobian. 
The work in \cite{sqp} redefines (\ref{eq:ik1}) as a quadratic program in the form of (\ref{eq:qp0}), which is iterated to find a solution. This quadratic programme will find the minimum-norm solution for $\dvec{q}$ at each step, which will be the same solution as given by (\ref{eq:ik1}).

A naive approach to joint-limit avoidance is to perform a global search (as explained in Part 1) while discarding solutions that exceed iteration limits and joint limits. Alternatively, the popular KDL inverse kinematics solver will not allow joint limits to be exceeded during iteration -- they are clamped to the joint limits \cite{kdl}. 

We can greatly improve the solvability by using the null-space projection devised by Baur \cite{gpm2}
\begin{align} \label{eq:ikgp1}
    \vec{q}_{k+1} &=
    \vec{q}_k +
    {\mat{J}(\vec{q}_k)}^{+} \vec{e}_k +
    \vec{q}_{null} \\
    \vec{q}_{null} &= \label{eq:qnull}
    \Big(
        \left(
            \mat{1}_n - \mat{J}(\vec{q})^{+} \mat{J}(\vec{q})
        \right)
        \Big(
            \frac{1}{\lambda_\Sigma}\vec{\Sigma}
        \Big)
    \Big) \\
    \vec{\Sigma} &=
    \left\{
        \begin{matrix*}[l]
            \sum_{i=1}^n
            \dfrac{(q_i - \bar{q}_{M_i})^2}
                  {(q_{M_i} - \bar{q}_{M_i})^2}
            & q_i \geq \bar{q}_{M_i} \\
            \sum_{i=1}^n
            \dfrac{(q_i - \bar{q}_{m_i})^2}
                  {(q_{m_i} - \bar{q}_{m_i})^2}
            & q_i \leq \bar{q}_{m_i} \\
            0 & \mbox{otherwise}
        \end{matrix*}
    \right.
\end{align}
where the maximum and minimum joint angles are specified by $q_{M_i}$ and $q_{m_i}$ respectively, while the maximum and minimum joint angle threshold are specified by $\bar{q}_{M_i}$ and $\bar{q}_{m_i}$. Once this threshold is passed, further progress toward the joint limit is penalised by $\vec{\Sigma}$. While this addition can help avoid joint limits, it does not guarantee joint-limit avoidance. The term $\lambda_\Sigma \in \mathbb{R}^+$ is a gain that adjusts how aggressively the joint limit is avoided. Furthermore, we can add the manipulability Jacobian to the null-space term as we did in (\ref{eq:ns1})
\begin{align}
\vec{q}_{null} &= \label{eq:qjnull}
\Big(
    \left(
        \mat{1}_n - \mat{J}(\vec{q})^{+} \mat{J}(\vec{q})
    \right)
    \Big(
        \frac{1}{\lambda_\Sigma}\vec{\Sigma} + \frac{1}{\lambda_m} \mat{J}_m(\vec{q})
    \Big)
\Big)
\end{align}
where $\lambda_m \in \mathbb{R}^+$ is a gain that adjusts how aggressively the manipulability is to be maximised.

\newcolumntype{M}[1]{>{\centering\arraybackslash}m{#1}}

\begin{table*}[pb!]
    \centering
    \small
    \renewcommand{\arraystretch}{1.3}

    \caption{Numerical IK Methods Compared over $10\,000$ Problems on a 6 DoF UR5 Manipulator}
    \label{tab:ik1}

    \begin{tabular}{ m{4.2cm} | M{0.9cm} | M{1.0cm}  | M{1.3cm} |M{1.4cm} | M{1.4cm} | M{1.5cm} | M{0.7cm} | M{1.2cm} }
    \hline
    \centering{Method} & Mean Iter. & Median Iter. & Infeasible Count & Mean Searches & Max Searches & Joint Limit Violations & Time per Iter. & Median Time\\
    \hline\hline
    NR                &  27.96  &  16.0  &  0  &  1.44  &  25.0  &  0  &  1.68  &   26.9  \\
    LM (Chan)         &  15.52  &   8.0  &  0  &  1.21  &  14.0  &  0  &  1.04  &   8.35  \\
    \hline
    LM$^+$ (Wampler)  &  23.75  &  13.0  &  0  &  1.35  &  20.0  &  0  &   1.0  &   13.0  \\
    LM$^+$ (Chan)     &  15.52  &   8.0  &  0  &  1.21  &  14.0  &  0  &  1.04  &   8.29  \\
    LM$^+$ (Sugihara) &  21.89  &  13.0  &  0  &  1.27  &  19.0  &  0  &  1.07  &  13.97  \\
    \hline
    QP $(\vec{J}_m)$  &  15.93  &   8.0  &  0  &  1.22  &  13.0  &  0  &  3.12  &  24.99  \\
    \end{tabular}
\end{table*}

\begin{table*}[pb!]
    \centering
    \small
    \renewcommand{\arraystretch}{1.3}

    \caption{Numerical IK Methods Compared over $10\,000$ Problems on a 7 DoF Panda Manipulator}
    \label{tab:ik2}

    \begin{tabular}{ m{4.2cm} | M{0.9cm} | M{1.0cm}  | M{1.3cm} | M{1.4cm} | M{1.4cm} | M{1.5cm} | M{0.7cm} | M{1.2cm} }
    \hline
    \centering{Method} & Mean Iter. & Median Iter. & Infeasible Count & Mean Searches & Max Searches & Joint Limit Violations & Time per Iter. & Median Time\\
    \hline\hline
    NR                                                          &   27.88  &   16.0  &    0  &   1.43  &   12.0  &  6705  &  1.71  &   27.35 \\
    LM (Chan)                                                   &   11.91  &    8.0  &    0  &   1.12  &    7.0  &  5394  &  1.06  &    8.46 \\
    \hline
    NR$^+$                                                      &  139.56  &   80.0  &  104  &    7.5  &  100.0  &     0  &  1.54  &  123.03 \\
    LM$^+$ (Wampler)                                            &  127.61  &   76.0  &  102  &   7.11  &   98.0  &     0  &   1.0  &    76.0 \\
    LM$^+$ (Chan)                                               &   37.55  &   18.0  &   91  &   3.81  &   86.0  &     0  &  1.03  &   18.54 \\
    LM$^+$ (Sugihara)                                           &   50.13  &   26.0  &   89  &   3.64  &   76.0  &     0  &  1.07  &   27.81 \\
    \hline
    NR$^+$ $\vec{q}_{null}(\vec{\Sigma})$                       &   347.7  &  219.0  &  254  &  15.88  &   99.0  &     0  &  3.02  &  661.59 \\
    LM$^+$ (Wampler) $\vec{q}_{null}(\vec{\Sigma})$             &  353.84  &  196.0  &  190  &  14.02  &   99.0  &     0  &  2.67  &  523.24 \\
    LM$^+$ (Chan) $\vec{q}_{null}(\vec{\Sigma})$                &    37.4  &   18.0  &   91  &   3.79  &   86.0  &     0  &  2.73  &   49.16 \\
    LM$^+$ (Sugihara) $\vec{q}_{null}(\vec{\Sigma})$            &   44.63  &   24.0  &   99  &   2.85  &   97.0  &     0  &  2.77  &   66.43 \\
    \hline
    NR$^+$ $\vec{q}_{null}(\vec{\Sigma}, \vec{J}_m)$            &  232.16  &  132.0  &  135  &  10.19  &   99.0  &     0  &  4.68  &  618.41 \\
    LM$^+$ (Wampler) $\vec{q}_{null}(\vec{\Sigma}, \vec{J}_m)$  &  178.22  &  103.0  &  105  &   8.58  &  100.0  &     0  &  4.23  &  435.74 \\
    LM$^+$ (Chan) $\vec{q}_{null}(\vec{\Sigma}, \vec{J}_m)$     &   37.33  &   18.0  &   90  &   3.77  &   86.0  &     0  &  4.25  &   76.53 \\
    LM$^+$ (Sugihara) $\vec{q}_{null}(\vec{\Sigma}, \vec{J}_m)$ &   49.55  &   26.0  &   89  &    3.6  &   97.0  &     0  &  4.29  &   111.5 \\
    \hline
    QP $(\vec{\Sigma}, \vec{J}_m)$                              &   42.42  &   14.0  &   76  &   2.12  &   86.0  &     0  &  3.95  &   55.36 \\
    \end{tabular}
\end{table*}

\begin{table*}[pb!]
    \centering
    \small
    \renewcommand{\arraystretch}{1.3}

    \caption{Numerical IK Methods Compared over $10\,000$ Problems on a 13 DoF (waist, arm \& index finger) Valkyrie Humanoid}
    \label{tab:ik3}

    \begin{tabular}{ m{4.2cm} | M{0.9cm} | M{1.0cm}  | M{1.3cm} |M{1.4cm} | M{1.4cm} | M{1.5cm} | M{0.7cm} | M{1.2cm} }
    \hline
    \centering{Method} & Mean Iter. & Median Iter. & Infeasible Count & Mean Searches & Max Searches & Joint Limit Violations & Time per Iter. & Median Time\\
    \hline\hline
    LM (Chan)                                                   &    6.31  &    6.0  &     0  &    1.0  &    1.0  &  9542  &  1.28  &    7.69  \\
    \hline
    NR$^+$                                                      &  285.88  &  235.0  &  2791  &  34.57  &  100.0  &     0  &  1.38  &  323.14  \\
    LM$^+$ (Chan)                                               &  156.13  &   98.0  &  1765  &  25.22  &  100.0  &     0  &   1.0  &    98.0  \\
    \hline
    NR$^+$ $\vec{q}_{null}(\vec{\Sigma})$                       &    82.6  &   37.0  &   109  &   6.82  &  100.0  &     0  &  2.97  &  109.88  \\
    LM$^+$ (Wampler) $\vec{q}_{null}(\vec{\Sigma})$             &   82.43  &   37.0  &   109  &    6.8  &  100.0  &     0  &  2.73  &  101.01  \\
    LM$^+$ (Chan) $\vec{q}_{null}(\vec{\Sigma})$                &   28.18  &   15.0  &    56  &   2.11  &   95.0  &     0  &  2.87  &   43.03  \\
    LM$^+$ (Sugihara) $\vec{q}_{null}(\vec{\Sigma})$            &   25.59  &   13.0  &    50  &   1.79  &  100.0  &     0  &  2.93  &   38.15  \\
    \hline
    LM$^+$ (Chan) $\vec{q}_{null}(\vec{\Sigma}, \vec{J}_m)$     &   28.69  &   15.0  &    60  &   2.14  &   99.0  &     0  &  4.63  &    69.4  \\
    LM$^+$ (Sugihara) $\vec{q}_{null}(\vec{\Sigma}, \vec{J}_m)$ &   24.86  &   13.0  &    56  &   1.74  &   91.0  &     0  &   4.7  &   61.16  \\
    \hline
    QP $(\vec{\Sigma}, \vec{J}_m)$                              &   15.29  &    7.0  &     0  &   1.27  &   18.0  &     0  &  4.19  &    29.3  \\
    \end{tabular}
\end{table*}

In Part 1, we showed that the Levenberg-Marquardt (LM) method
\begin{align} \label{eg:aik0}
    \vec{q}_{k+1} 
    &= 
    \vec{q}_k +
    \left(
        \mat{A}_k
    \right)^{-1}
    \vec{g}_k \\
    \mat{A}_k
    &=
    {\mat{J}(\vec{q}_k)}^\top
    \mat{W}_e \
    {\mat{J}(\vec{q}_k)}
    +
    \mat{W}_n \\
    \vec{g}_k &=
    {\mat{J}(\vec{q}_k)}^\top
    \mat{W}_e
    \vec{e}_k
\end{align}
provided much better results for inverse kinematics than the NR method,
where $\mat{W}_n = \mbox{diag}(\vec{w_n})\big(\vec{w_n} \in (\mathbb{R}^+)^n \big)$ is a diagonal damping matrix. In Part 1, we detailed the choice of $\mat{W}_n$ based on proposals by Wampler \cite{wampler}, Chan \cite{chan}, and Sugihara \cite{ik3}.

As with the NR method, we can naively perform joint limit avoidance with the LM method through global search, and discard solutions that exceed iteration limits and joint limits.
We can improve this approach by augmenting the vector $\vec{g}_k$ with the same null-space motion defined in (\ref{eq:qnull})
\begin{align}
\vec{g}_k &=
{\mat{J}(\vec{q}_k)}^\top
\mat{W}_e
\vec{e}_k
+ \vec{q}_{null}.
\end{align}

The addition of null-space motion provides much better results for inverse kinematics when trying to avoid joint limits, but it is only available on redundant robots. For improved constrained inverse kinematics, we can use the augmented QP with slack from (\ref{eq:mmc}) with
\begin{align}
    \vec{q}_{k+1} = \vec{q}_{k} + \dvec{q}.
\end{align}

To increase the likelihood of finding an IK solution, the TRAC-IK algorithm \cite{tracik}, which is used as the default IK solver for the popular robotics software package Moveit \cite{moveit}, runs two IK solvers in parallel. The first is the NR method shown in (\ref{eq:ik1}), and the second is an NR method redefined as a quadratic programme with a custom error metric.

\subsection*{Inverse Kinematics Comparison}

We show a comprehensive comparison of various numerical inverse kinematics solvers on three different types of robots. For each robot, the IK algorithm attempts to reach $10\,000$ randomly generated valid end-effector poses and results are summarised in Tables \ref{tab:ik1}-\ref{tab:ik3}.
All methods use a global search with a 30 iteration limit within a search, and a maximum of 100 searches. The Infeasible Count column reports how many solutions failed to converge after 100 searches -- zero is best.
For the LM Wampler method, we use $\lambda=1e-4$. For the LM Sugihara method, we use $w_n = 0.001$. Methods with a $(\cdot)^+$ indicate that solutions with a joint-limit violation are treated as a failure and another attempt is performed. Methods marked with $\vec{q}_{null}(\vec{\Sigma})$ have joint limit avoidance projected into the null-space using (\ref{eq:qnull}). Methods marked with $\vec{q}_{null}(\vec{\Sigma}, \mat{J}_m)$ have joint limit avoidance and manipulability maximisation projected into the null-space using (\ref{eq:qjnull}). For the QP method, the symbols $\vec{\Sigma}$ and $\mat{J}_m$ indicate joint limit avoidance and manipulability maximisation have been incorporated respectively.
Note that null-space methods can not be used on the UR5 manipulator as it is not redundant. The time per iteration represents the average of how long one iteration of the corresponding method took relative to the fastest method iteration within the Table. The rightmost column is the relative number of iterations to find a solution based on the time per iteration and the median number of iterations.

Table \ref{tab:ik1} displays the results on a six degree-of-freedom UR5 manipulator where each joint can articulate $\pm 180^\circ$ -- this means that for each joint, every angle is achievable, which is a larger range than most other manipulators.
Table \ref{tab:ik2} displays the results on a seven degree-of-freedom Panda manipulator where each joint has unique articulation limits, more typical of other manipulators.
Table \ref{tab:ik3} displays the results on a thirteen degree-of-freedom kinematic chain within the Valkyrie humanoid robot involving the waist, right shoulder, right arm, and right index finger. Several joint limits within this kinematic chain are quite small, with some having a total range of $20^\circ$.

This experiment, running on non-redundant, redundant, and hyper-redundant robots with wide, medium, and narrow joint limits respectively, exposes the key differences between each IK algorithm along with some strengths and weaknesses.
The first section of Tables \ref{tab:ik2} and \ref{tab:ik3}, on robots with joints that have unachievable coordinates, shows many solutions with joint limit violations. This shows that, when not constrained otherwise, the IK solvers will converge on a solution that violates the joint limits of the robot the majority of the time. 
The Chan, Wampler, and Sugihara solvers are consistently the fastest per step. The time cost increases as we add extra functionality, such as active joint-limit avoidance and manipulability maximisation. The QP IK solver is clearly the most robust, uses the least iterations in the median, and will use fewer attempts to find a solution, but is one of the slowest algorithms presented. As more degrees-of-freedom are added and the joint range becomes narrower, the QP solver improves in solution speed relative to the other solvers.

On the UR5, where every joint angle is achievable, Chan's method without any null-space terms is both the fastest and most reliable.
Interestingly, the base Chan and Sugihara methods on the Panda slightly outperform the Chan and Sugihara methods with joint limit avoidance, and manipulability maximisation. Although the latter methods will typically arrive at a solution in fewer iterations, the extra computation time makes the median time much worse than the base methods. On the Panda, both the base Chan and the QP solver provide the best overall results. On the Valkyrie, the QP solver clearly outperforms the other solvers, even when accounting for the additional time per step.


\section*{Singularity Escapability}

A robot can lose degrees of freedom if a joint is at its limit, the arm is fully extended or when two or more joint axes align. In the last failure case, the manipulator is in a singular configuration.

At a singularity, the manipulator Jacobian becomes rank deficient, and when approaching a singularity, the Jacobian becomes ill-conditioned. For Jacobian-based motion controllers, the demanded joint velocities calculated from the Jacobian inverse will approach infinity as the singularity is approached. The Moore-Penrose pseudoinverse is a common approach to avoiding this issue, however the performance is not reliable in all cases. There have been several works that, using the manipulator Jacobian and Hessian, can determine if a singularity is escapable, and if so, which joints should be actuated to do so \cite{se2, se3, se4}. In the remainder of this tutorial, we detail a quadratic-rate motion control that can control a robot when away from, near to, or even at a singularity.

\subsection*{Quadratic-Rate Motion Control}

Quadratic-rate motion control is a method of controlling a manipulator through or near a singularity \cite{se1}. By using the second-order differential kinematics, the controller does not break down at or near a singularity, as the resolved-rate motion controller would. We start by considering the end-effector pose $\vec{x} = f(\vec{q}) \in \mathbb{R}^6$ given by the forward kinematics.
Introducing a small change to the joint coordinates $\Delta\vec{q}$, we can write
\begin{align} \label{eq:qr1}
    \vec{x} + \Delta\vec{x} &= f(\vec{q} + \Delta\vec{q}).
\end{align}
and the Taylor series expansion is
\begin{align}
    \vec{x} + \Delta\vec{x}
    &= f(\vec{q}) +
    \dfrac{\partial f}{\partial \vec{q}} \Delta\vec{q} +
    \dfrac{1}{2}
    \left(
        \dfrac{\partial^2 f}{\partial \vec{q}^2} \Delta\vec{q}
    \right) \Delta\vec{q} +
    \hdots \nonumber \\
    \Delta\vec{x}
    &=
    \mat{J}(\vec{q}) \Delta\vec{q} +
    \frac{1}{2}
    \Big(
        \mat{H}(\vec{q}) \Delta\vec{q}
    \Big) \Delta\vec{q} +
    \hdots \label{eq:qr2}
\end{align}
where for quadratic-rate control we wish to retain both the linear and quadratic terms (the first two terms) of the expansion.

To form our controller, we use an iteration-based Newton-Raphson approach to solve for $\Delta\vec{q}$ in (\ref{eq:qr2}).
Firstly, we rearrange (\ref{eq:qr2}) as
\begin{align}
    \vec{g}(\Delta\vec{q}_k)
    &=
    \mat{J}(\vec{q}_k) \Delta\vec{q}_k +
    \frac{1}{2}
    \Big(
        \mat{H}(\vec{q}_k) \Delta\vec{q}_k
    \Big) \Delta\vec{q}_k -
    \Delta\vec{x} = 0 
\end{align}
where $\vec{q}_k$ and $\Delta\vec{q}_k$ are the manipulator's current joint coordinates and change in joint coordinates respectively, and $\Delta\vec{x}$ is the desired change in end-effector position. Taking the derivative of $\vec{g}$ with respect to $\Delta\vec{q}$
\begin{align}
    \dfrac{\partial \vec{g}(\Delta\vec{q}_k)}
          {\partial \Delta\vec{q}}
    &=
    \mat{J}(\vec{q}_k) + \mat{H}(\vec{q}_k)\Delta\vec{q}_k \nonumber \\
    &= \emat{J}(\vec{q}_k, \Delta\vec{q}_k)
\end{align}
we obtain a new Jacobian $\emat{J}(\vec{q}_k, \Delta\vec{q}_k)$. Using this Jacobian we can create a new linear system
\begin{align}
    \emat{J}(\vec{q}_k, \Delta\vec{q}_k) \vec{\delta}_{\Delta\vec{q}} &= \vec{g}(\Delta\vec{q}_k) \nonumber \\
    \vec{\delta}_{\Delta\vec{q}} &= \emat{J}(\vec{q}_k, \Delta\vec{q}_k)^{-1} \vec{g}(\Delta\vec{q}_k)
\end{align}
where $\vec{\delta}_{\Delta\vec{q}}$ is the update to $\Delta\vec{q}_k$. The change in joint coordinates at the next step are
\begin{align}
    \Delta\vec{q}_{k+1} = \Delta\vec{q}_{k} - \vec{\delta}_{\Delta\vec{q}.}
\end{align}

In the case that $\Delta\vec{q} = \vec{0}$, quadratic-rate control reduces to resolved-rate control. This is not suitable if the robot is in or near a singularity. Therefore, a value near $0$ may be used to seed the initial value with $\Delta\vec{q} = \left(0.1,\ \cdots,\ 0.1 \right)$, or the pseudo-inverse approach could be used
\begin{align}
    \Delta\vec{q} = \mat{J}(\vec{q})^{+} \Delta\vec{x}.
\end{align}

This controller can be used in the same manner as resolved-rate motion control described in Part 1. The manipulator Jacobian and Hessian are calculated in the robot's base frame, and the end-effector velocity $\Delta\vec{x}$ can be calculated using the angle-axis method described in Part 1.

As quadratic-rate motion control is another form of advanced velocity control, many of the techniques described in the Advanced Velocity Control section can also be applied here. It may also be adapted into an inverse kinematics algorithm using any technique described in the Inverse Kinematics section in Part 1, or the Advanced Inverse Kinematics section in this article.

In the literature \cite{se1, se2, se3, se4}, the manipulator Hessian is reported as taking too long to compute to be used in closed-loop control. However, with the approach detailed in this article and the implementation provided by the Robotics Toolbox for Python, the manipulator Hessian can be computed in less than 1\unit{\mu s} \cite{rtb}. Therefore, in the current day, that argument is no longer valid.


\section*{Acknowledgments}

We are grateful to the anonymous reviewers whose detailed and insightful comments have improved this article.
This research was conducted by the Australian Research Council project number CE140100016, and supported by the QUT Centre for Robotics (QCR). We would also like to thank the members of QCR who provided valuable feedback and insights while testing this tutorial and associated Jupyter Notebooks.


\section*{Conclusion}

In this tutorial's final instalment, we have covered many advanced aspects of manipulator differential kinematics. We first described a procedure for describing any manipulator's second-and higher-order differential kinematics.
We then detailed how differential kinematics can be translated into analytical forms, which is highly advantageous for task-space dynamics applications.
We detailed how numerical inverse kinematics solvers can be extended to avoid joint limits and maximise manipulability. Finally, we described how the manipulator Hessian is used to create quadratic-rate motion control, a controller which will work when the robot is in, or near, a singularity.

This tutorial is not exhaustive, and many important and useful techniques have not been visited. For example, resolved-acceleration control can be used for trajectory following, where the required velocities and accelerations are known beforehand. Additionally, task-space control approaches such as operational-space control make strong use of differential kinematics.


{\footnotesize
\printbibliography[title={References}]}


\end{document}